\documentclass[letterpaper]{article} % DO NOT CHANGE THIS
\usepackage{aaai2026}  % DO NOT CHANGE THIS
\usepackage{times}  % DO NOT CHANGE THIS
\usepackage{helvet}  % DO NOT CHANGE THIS
\usepackage{courier}  % DO NOT CHANGE THIS
\usepackage[hyphens]{url}  % DO NOT CHANGE THIS
\usepackage{graphicx} % DO NOT CHANGE THIS
\usepackage{booktabs}
\usepackage{amsfonts}
\usepackage{amsmath}
\usepackage{nccmath}
\usepackage{xcolor}
\usepackage{colortbl}
\usepackage{siunitx} 
\usepackage{subcaption}\usepackage{subcaption}
\usepackage[numbers]{natbib}  % DO NOT CHANGE THIS AND DO NOT ADD ANY OPTIONS TO IT
\usepackage{caption} % DO NOT CHANGE THIS AND DO NOT ADD ANY OPTIONS TO IT
\definecolor{myyellow}{RGB}{255,242,204} % 浅黄色
\definecolor{myblue}{RGB}{222,235,247}   % 浅蓝色
\definecolor{mygray}{RGB}{251,229,214}   % 浅灰色
\usepackage{algorithm}
\usepackage{algorithmic}

\usepackage{newfloat}
\usepackage{listings}
\DeclareCaptionStyle{ruled}{labelfont=normalfont,labelsep=colon,strut=off} % DO NOT CHANGE THIS
\floatstyle{ruled}
\newfloat{listing}{tb}{lst}{}
\floatname{listing}{Listing}
\title{\emph{Emotion-o1}: Adaptive Long Reasoning for Emotion Understanding in LLMs}
\author{
    Changhao Song,
    Yazhou Zhang\textsuperscript{†},
    Hui Gao,
    Kaiyun Huang,
    Peng Zhang\textsuperscript{‡}
}
\affiliations{
    College of Intelligence and Computing, Tianjin University, Tianjin, China \\
    \textsuperscript{†}yzhou\_zhang@tju.edu.cn, 
    \textsuperscript{‡}pzhang@tju.edu.cn, \\
    songchanghao@tju.edu.cn, hui\_gao@tju.edu.cn, hky666888@tju.edu.cn
}

\usepackage{bibentry}
\begin{document}

\maketitle

\begin{abstract}
Long chain-of-thought (CoT) reasoning has shown great promise in enhancing the emotion understanding performance of large language models (LLMs). However, current fixed-length CoT methods struggle to balance reasoning depth and efficiency. Simple tasks (e.g., sentiment classification) are over-reasoned, while complex tasks (e.g., sarcasm understanding) lack depth. To fill this gap, we present Emotion-o1, an adaptive CoT framework that dynamically adjusts reasoning length based on emotion-task complexity. Emotion-o1 is trained by distilling adaptive CoT patterns from a reasoning-oriented LLM, followed by supervised fine-tuning and reinforcement learning with a four-part reward targeting accuracy, brevity, structure, and redundancy. Experimental results on four emotion tasks highlight: (1) Emotion-o1 demonstrates significant improvements over its backbone, with F1 score increases of 10\%↑(Sentiment), 5\%↑(Emotion), 18\%↑(Humor), and 27\%↑(Sarcasm). (2) In sentiment and sarcasm tasks, our 8B model demonstrates superior performance against advanced LLMs, outperforming Grok-3 by 1.1\% and Claude-3.7 by 2\%. (3) The framework maintains accuracy while reducing reasoning length by 83\% compared to OpenAI-o1, demonstrating effective precision-efficiency optimization. Emotion-o1 effectively balances reasoning depth and efficiency for emotion understanding in LLMs.

% The emergence of large language models (LLMs) has significantly advanced the performance in emotion recognition tasks. However, current approaches typically employ fixed-length chain-of-thought (CoT) reasoning, limiting their adaptability to the inherent complexity and variability of emotional analysis. To address this limitation, we propose a length-adaptive reasoning framework that distills variable-length reasoning chains from reasoning-oriented LLMs with strong inference capabilities, integrating task-specific fine-tuning to generate flexible reasoning paths. Furthermore, we integrate reinforcement learning (RL) to design a comprehensive reward function that balances multiple optimization objectives, including prediction accuracy, depth adaptation, structural diversity, and redundancy suppression. Experimental evaluations demonstrate that our proposed method achieves reasoning performance comparable to state-of-the-art LLMs across multiple emotion recognition tasks, yielding significant improvements of \textbf{6\%–27\%} in Acc and F1 scores. Overall, our work effectively bridges rigid cognitive-theoretic reasoning and the nuanced complexity of emotion recognition through adaptive depth analysis.
\end{abstract}


\section{Introduction}

% This document details the formatting requirements for anonymous submissions. The requirements are the same as for camera ready papers but with a few notable differences:

% \begin{itemize}
%     \item Anonymous submissions must not include the author names and affiliations. Write ``Anonymous Submission'' as the ``sole author'' and leave the affiliations empty.
%     \item The PDF document's metadata should be cleared with a metadata-cleaning tool before submitting it. This is to prevent leaked information from revealing your identity.
%     \item References must be anonymized whenever the reader can infer that they are to the authors' previous work.
%     \item AAAI's copyright notice should not be included as a footer in the first page.
%     \item Only the PDF version is required at this stage. No source versions will be requested, nor any copyright transfer form.
% \end{itemize}

% You can remove the copyright notice and ensure that your names aren't shown by including \texttt{submission} option when loading the \texttt{aaai2026} package:

% \begin{quote}\begin{scriptsize}\begin{verbatim}
% \documentclass[letterpaper]{article}
% \usepackage[submission]{aaai2026}
% \end{verbatim}\end{scriptsize}\end{quote}

% The remainder of this document are the original camera-
% ready instructions. Any contradiction of the above points
% ought to be ignored while preparing anonymous submis-
% sions.

\begin{figure}[t]
    \centering
    \includegraphics[width=1.05\columnwidth]{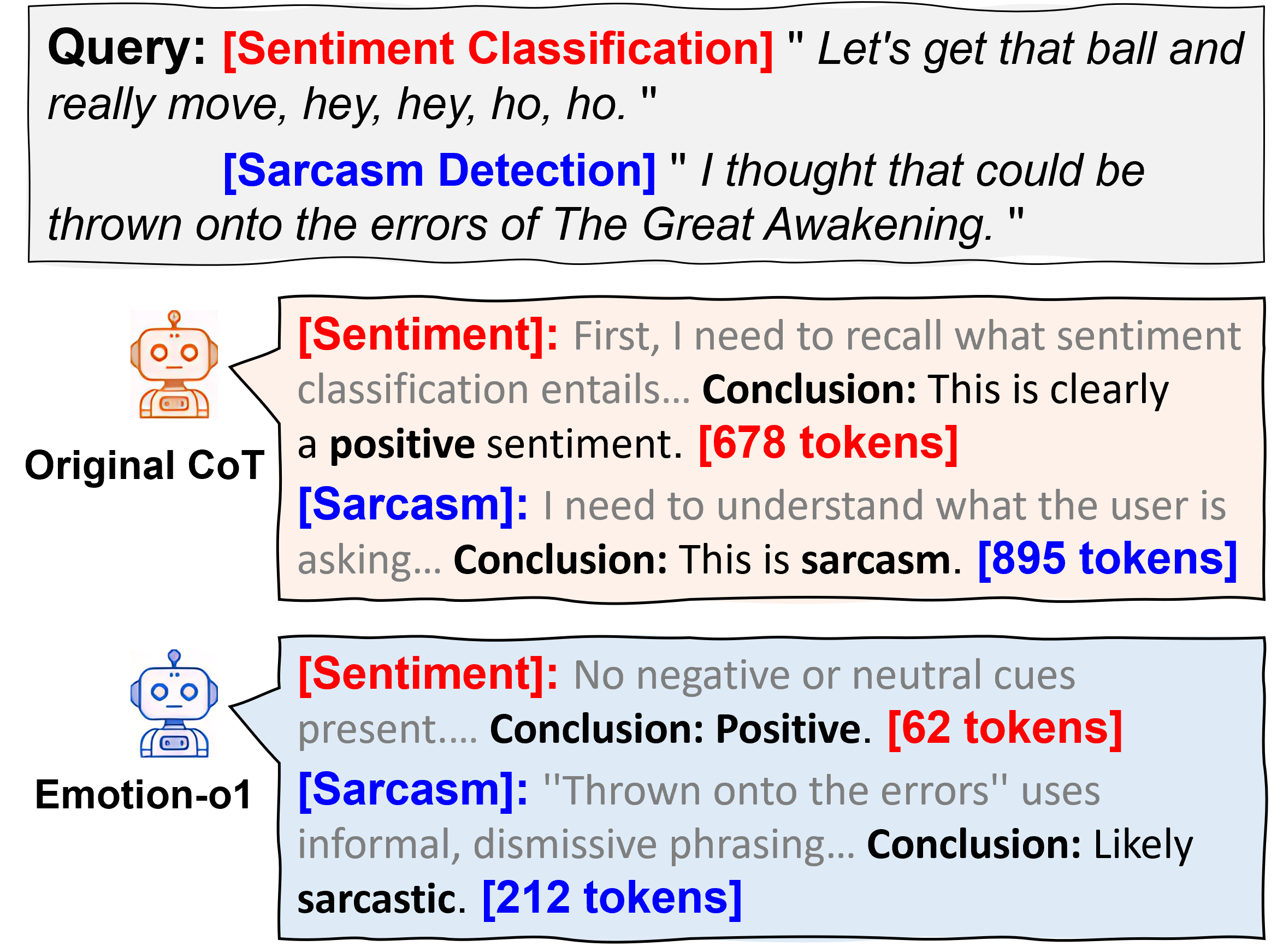}
    \caption{Original long CoT may lead to redundant computations or insufficient reasoning against our Emotion-o1.}
    \label{fig:motivation}
\end{figure}

CoT reasoning, which elaborates a series of intermediate steps, has significantly improved the ability of LLMs to solve complex problems~\cite{yao2025sarcasm}. This has led to the rise of a new class of models known as large reasoning models (LRMs), such as DeepSeek-R1~\cite{deepseekai2025deepseekr1incentivizingreasoningcapability}, OpenAI-o1~\cite{openai2024openaio1card}, and Qwen-QwQ~\cite{qwq32b}. Such LRMs demonstrate that scaling CoT length to hundreds or even thousands of steps can yield continual gains in reasoning accuracy, interpretability, and robustness across a wide range of tasks.

Despite these advances, fixed-length CoT strategies are poorly suited for emotion understanding tasks. For instance, simple tasks such as binary sentiment classification (e.g., ``Is this review positive or negative?''), often elicit excessively verbose reasoning, resulting in substantial computational overhead and inefficient overthinking~\cite{xia2025tokenskipcontrollablechainofthoughtcompression}. In contrast, complex tasks such as sarcasm detection suffer from shallow reasoning, failing to capture nuanced pragmatic and contextual cues, as shown in Fig.~\ref{fig:motivation}. This disconnect between fixed reasoning lengths and the inherently dynamic nature of emotion understanding leads to both performance bottlenecks and wasted computation.

We posit that effective emotion reasoning demands adaptive flexibility. Simple emotion tasks benefit from short, efficient reasoning paths, while complex emotional phenomena such as irony, ambiguity, and humor require deeper, reflective chains of thought. 
However, existing CoT-based emotion understanding approaches lack the ability to dynamically adjust the length of the reasoning according to the complexity of the task, limiting their generalization across different emotion domains.

To fill this gap, we introduce \textbf{Emotion-o1}, an adaptive reasoning framework that dynamically adjusts CoT length according to the complexity of the emotional task.  Specifically, our approach first distills variable-length, structurally diverse reasoning paths, such as backtracking and self-reflection, etc., from SoTA LRMs (e.g., DeepSeek-R1). After supervised fine-tuning the model to acquire comprehensive reasoning capabilities, we further optimize reasoning quality via reinforcement learning, guided by a multi-objective reward function across four dimensions: prediction accuracy, depth adaptability, structural diversity, and redundancy suppression. This allows Emotion-o1 to develop emotionally aligned, length-adaptive reasoning strategies tailored to the demands of each task.

Given that sentiment classification and emotion recognition mainly involve shallow emotional cues and limited contextual dependencies, we follow prior work in treating them as basic tasks~\cite{evans2002emotion}. In contrast, sarcasm detection and humor understanding require complex pragmatic reasoning and deep contextual integration, and are therefore regarded as complex tasks that require deeper reasoning~\cite{chauhan2020sentiment,chauhan2022sentiment}.
We present empirical evaluations of the proposed approach on four emotion understanding tasks, and compare its performance against nine SoTA LLMs (e.g., DeepSeek-R1, GPT-4o, Claude 3.7, etc.).
We highlight three key findings: (1) compared to the backbone, Emotion-o1 achieves F1-score improvements of 10\%, 5\%, 18\%, and 27\% on the four tasks, demonstrating the effectiveness of incorporating diverse reasoning structures; (2) Emotion-o1 achieves SoTA performance in sentiment classification. In sarcasm recognition, its F1 score is only 1\% lower than that of GPT-4o, indicating our 8B model competitively matches leading LLMs; (3)  compared to OpenAI-o1 (DeepSeek-R1), Emotion-o1 reduces the average reasoning length by 73\% (54\%), 52\% (27\%), 83\% (70\%), and 70\% (58\%) across the four tasks, highlighting its efficiency advantage.
Our main contributions are as follows:
\begin{itemize}
    \item We propose \textbf{Emotion-o1}, an adaptive CoT reasoning framework that dynamically adjusts reasoning length based on the complexity of emotion understanding tasks.
    
    \item We design a multi-objective reward function that jointly optimizes for prediction accuracy, reasoning brevity, structural coherence, and redundancy suppression, enabling the LRM to learn emotionally aligned and task-adaptive reasoning strategies.
    
    \item We validate Emotion-o1 on four emotion tasks, achieving SoTA performance with notably reduced reasoning cost.
\end{itemize}

% Overall, our approach establishes an adaptive connection between structured reasoning and emotional cognition, enabling context-sensitive adjustments from shallow emotional analysis to deep pragmatic reasoning.

\section{Related Work}

\begin{figure*}[t]
\centering
\includegraphics[width=1.0\textwidth]{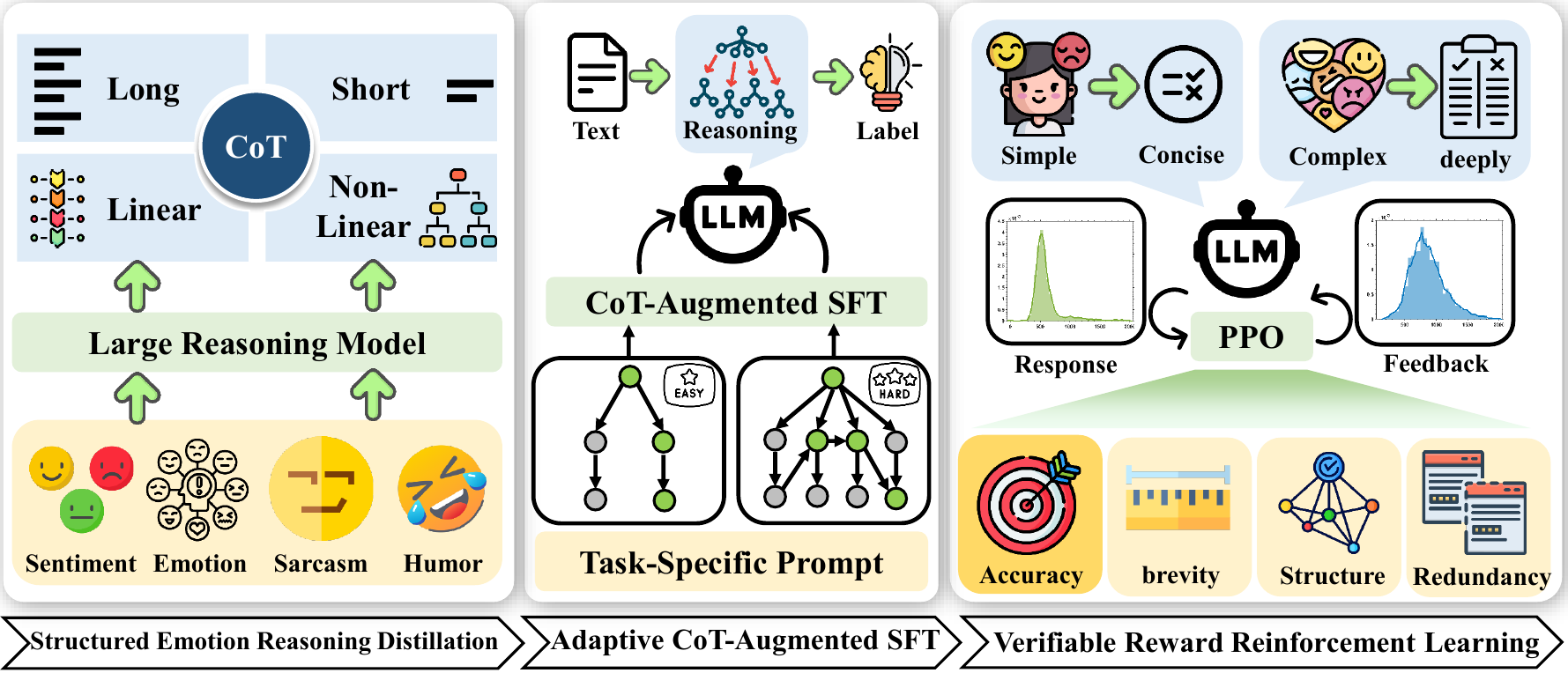}
\caption{Overview of the proposed framework.}
\label{fig:pipeline}
\end{figure*}

\subsection{Affective Computing}

Affective computing (AC) aims to endow machines with the ability to recognize, interpret, and respond to human emotions~\cite{zhang2023cmma}. Early works relied on feature engineering and traditional classifiers. With the rise of pre-trained language models (PLMs) such as BERT \cite{devlin2019bert}, fine-tuning on labeled datasets became the dominant approach for both affective understanding (AU) and affective generation (AG) \cite{verma2021techniques, nie2022review}. However, these methods struggled to generalize across domains and lacked flexibility for multi-task emotion reasoning~\cite{mao2022biases}.

LLMs brought a paradigm shift, enabling zero-shot and instruction-based emotion modeling \cite{brown2020language, zhou2022large}. While LLMs showed strong generalization in basic AU tasks, their performance remained limited in fine-grained, pragmatic tasks such as sarcasm and humor detection~\cite{zhang2024sarcasmbenchevaluatinglargelanguage}. Recent efforts introduced CoT prompting to improve reasoning, but existing methods typically adopt fixed-length templates and lack adaptability to emotional task complexity.

Our work departs from this trend by distilling variable-length, structure-rich reasoning traces from powerful models. Emotion-o1 integrates multi-stage training with a multi-objective reward to support dynamic reasoning depth.

\subsection{Chain-of-Thought Reasoning}
CoT reasoning has become a central technique for enhancing the reasoning capabilities of LLMs. Early research focused on short CoT, which involves shallow, linear reasoning paths with limited logical depth and little capacity for exploration or error correction \cite{chen2024unlocking}. However, these methods struggle with complex tasks that require revisiting earlier steps or considering alternative paths \cite{mirzadeh2024gsm}. 

To overcome these issues, recent efforts have explored non-linear reasoning structures, enabling richer inference dynamics. Frameworks such as Tree-of-Thoughts (ToT) \cite{yao2023tree} and Graph-of-Thoughts (GoT) \cite{besta2024graph} introduce branching and parallel reasoning, allowing models to maintain multiple hypotheses and selectively backtrack. These methods laid the groundwork for long CoT.

Now, long-CoT LLMs, such as OpenAI-O1 \cite{openai2024openaio1card} and DeepSeek-R1 \cite{deepseekai2025deepseekr1incentivizingreasoningcapability}, extend these ideas by scaling CoT to thousands of steps and incorporating dynamic feedback mechanisms. These systems achieve SoTA performance in mathematics, programming, and symbolic inference.

In contrast, our proposed Emotion-o1 bridges short and long CoT paradigms by dynamically adjusting reasoning depth and structure based on task complexity. It combines the efficiency of shallow reasoning with the flexibility and expressiveness of deep inference.

\section{Methodology}

As shown in Fig.~\ref{fig:pipeline}, our framework include three stages:  
(1) Structured Emotion Reasoning Distillation extracts variable-length reasoning paths from leading LRMs;  
(2) Adaptive CoT-Augmented SFT initializes the model with structured emotional reasoning ability;  
(3) Reward-based RL refines reasoning quality via multi-objective optimization.

\subsection{Task Definition}

Given an input text $x$, our goal is to model emotion understanding as a structured generation task, where the output sequence $y = [y_1, y_2, \ldots, y_T]$ consists of a reasoning path $\text{CoT}(x) = [y_1, \ldots, y_{T-1}]$ followed by a final prediction token $y_T$ representing the target label.

Formally, the model optimizes the conditional likelihood:
\[
P_\theta(y \mid x) = \prod_{t=1}^{T} P_\theta(y_t \mid x, y_{<t})
\]
where $\theta$ denotes model parameters and $y_{<t}$ refers to previously generated tokens.

The final token $y_T$ is deterministically mapped to a label $l \in \mathcal{L}$ via a mapping function $f_\text{map}$: $l = f_\text{map}(y_T)$, where $\mathcal{L}$ denotes the label space. 

Our objective is to jointly model the correctness of $l$ and the interpretability of $\text{CoT}(x)$, enabling adaptive reasoning aligned with task difficulty.

\subsection{Structured Emotion Reasoning Distillation}
\label{sec:Dataset}
We construct annotated samples with diverse reasoning paths by distilling a leading LRM.
Specifically, we select four representative emotion understanding tasks, each paired with a widely used benchmark dataset: MELD \cite{poria2018meld} for sentiment classification and emotion recognition, Sarcasm Corpus V2 \cite{oraby2017creating} for sarcasm detection, and Reddit Humor Detection \cite{humorDetection2019} for humor recognition. Each instance consists of a text input $x_i$ and its corresponding label $y_i$.

% To guarantee the quality of the CoT data, we employed widely acknowledged and representative datasets as standards for four typical emotion understanding tasks. Specifically, we chose the MELD\cite{poria2018meld} for sentiment and emotion classification, the Sarcasm Corpus V2\cite{oraby2017creating} for sarcasm detection, and the Reddit Humor Detection \cite{humorDetection2019} for humor detection. 

% Utilizing the ground-truth labels and texts from standard datasets, we leveraged carefully designed prompts to distill reasoning processes at various levels from the DeepSeek-R1. Among the generated CoTs, we identify and retain correct reasoning processes through a rejection sampling procedure, which are subsequently utilized for further training.

% We formalize the process of CoT distillation as follows. For each textual input-label pair $(x_i, y_i)$, we first construct a composite prompt $p(x_i, y_i, c)$, where $c$ specifies the reasoning strategy (linear or non-linear, short or long). Then we sample multiple candidate CoT trajectories from the LLM:
For each sample $(x_i, y_i)$, we construct a prompt template $p(x_i, y_i, c)$, where $c$ specifies the reasoning strategy, including \textit{structure type} (linear or non-linear) and \textit{length type} (short or long). We then use the DeepSeek-R1 for conditional sampling and generate $N$ candidate reasoning paths:
\begin{equation}
\{r_{i,j}\}_{j=1}^{N} \sim LLM(p(x_i, y_i, c))
\end{equation}
where $N$ is the number of candidate responses generated per prompt. We employ rejection sampling to filter responses, retaining only those matching the ground-truth label $y_i$:
\begin{equation}
\mathcal{R}_i = \{ r_{i,j} \mid label(r_{i,j}) = y_i, j \in [1, N] \}
\end{equation}
where $\mathcal{R}_i$ denotes the set of valid CoT reasoning responses.

\begin{figure}[t]
    \centering
    \includegraphics[width=1.0\columnwidth]{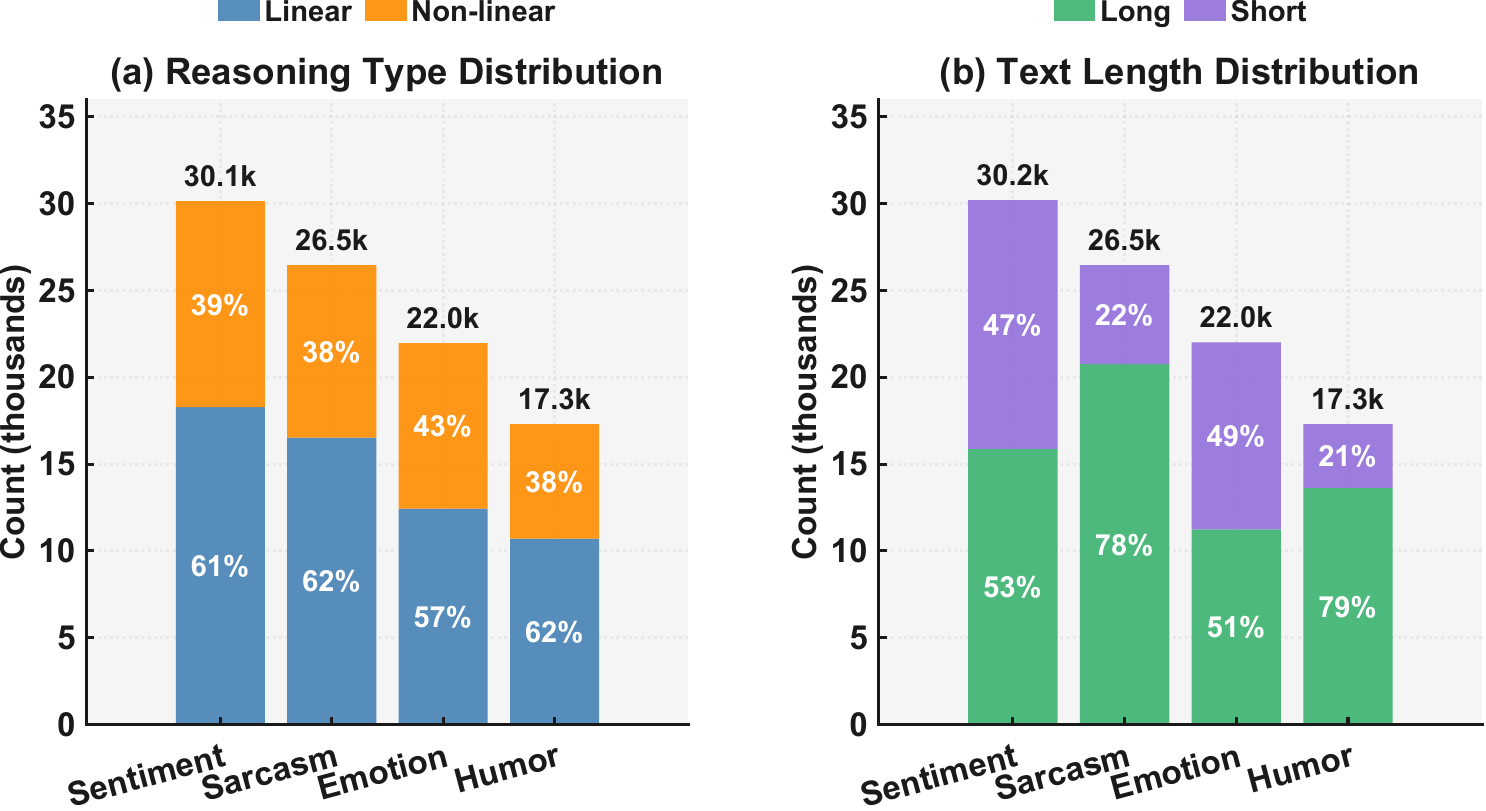}
    \caption{Overview of the CoT Dataset}
    \label{fig:dataset}
\end{figure}

Different reasoning strategies exhibit task-specific efficacy across emotion understanding tasks. Our prompt explicitly steers the model to generate responses with distinct reasoning strategies $c$. Specifically, we considered two primary dimensions of reasoning structure:
\begin{itemize}
\item \textbf{Length Type}: Short (concise, direct reasoning) and Long (comprehensive, detailed analysis).
\item \textbf{Reasoning Type}: Linear (step-by-step reasoning) and Non-linear (multi-path, branched reasoning structures such as Tree-of-Thoughts or Graph-of-Thoughts).
\end{itemize}
Thus, each textual input $x_i$ could yield multiple valid CoT responses across these dimensions, The final dataset $\mathcal{D}$ was constructed by aggregating all valid CoT reasoning instances across input samples and reasoning dimensions:
\begin{equation}
\medmath{
\mathcal{D} = \left\{ 
(x_i, y_i, r_{i,j}, c_{i,j}, l_{i,j}) 
\;\middle|\; 
\begin{Bmatrix}
r_{i,j} \in \mathcal{R}_i \\
c_{i,j} \in \{\text{linear}, \text{non-linear}\} \\
l_{i,j} \in \{\text{short}, \text{long}\}
\end{Bmatrix} 
\right\}
}
\end{equation}
The emotion CoT dataset is shown in Fig.~\ref{fig:dataset}. We compile \textbf{30K}, \textbf{22K}, \textbf{26K}, and \textbf{17K} samples for the \textit{sentiment}, \textit{emotion}, \textit{sarcasm}, and \textit{humor} tasks, respectively. After filtering, the ratio of linear to non-linear reasoning is approximately \textbf{6:4}, and that of long to short CoT is about \textbf{5:5} for Emotion and Sentiment tasks, and \textbf{8:2} for Humor and Sarcasm tasks.

\subsection{Adaptive CoT-Augmented SFT}

We propose an Adaptive CoT-Augmented SFT method to enhance the reasoning capabilities of LLMs for different emotion tasks. Given input text $x = {x_1, x_2, \dots, x_T}$ and corresponding CoT rationale $r$, we construct task-specific instruction prompts to guide structured reasoning during fine-tuning. The prompt construction is defined as:

\begin{equation}
\mathcal{P}(x, r, y, \mathcal{D}) = \Phi \oplus \Gamma(x) \oplus \Psi(r) \oplus \Omega(y, \mathcal{D})
\end{equation}
where $\Phi$ denotes the task identifier (e.g., `Emotion Classification Task`), $\Gamma(\cdot)$ formats the input text, $\Psi(\cdot)$ incorporates the reasoning steps, $\Omega(\cdot)$ encodes the label $y$ with class definitions from dataset $\mathcal{D}$, $\oplus$ represents string concatenation.

The training objective maximizes the conditional likelihood of the complete reasoning path and label prediction:
{\small
\begin{equation}
\mathcal{L}_{\text{SFT-CoT}} = -\sum_{t=1}^{L} \log P(w_t \mid \mathcal{P}(x, r, y, \mathcal{D}), w_{<t}; \theta)
\end{equation}
}
where $w_t$ denotes tokens in the prompt, $L$ is the response length, and $\theta$ represents model parameters. By dynamically incorporating task-specific classification schemas into the instructional prompt design, the model can flexibly adapt to diverse emotion taxonomies, thereby achieving robust generalization across multi-emotion tasks.

\subsection{Verifiable Reward RL}

We propose a verifiable reward RL approach to optimize the reasoning quality of the model further. Initialized with a fine-tuned SFT model, our method employs the proximal policy optimization (PPO) algorithm for training. By sampling multiple candidate responses during each update, the model learns to adjust its reasoning length according to task complexity adaptively.

The reward function is defined as a weighted combination of four components: prediction accuracy, depth adaptability, structural diversity, and redundancy suppression. Each component is formulated as follows.

\subsubsection{Accuracy Reward}

The first and most important reward is the accuracy reward, ensuring that the model prioritizes generating correct predictions aligned with the ground-truth labels. It is written as:
\begin{equation}
r_{\text{acc}} = 
\begin{cases}
+1.0, & \text{if} \quad \hat{y} = y \\
-1.0, & \text{if} \quad \hat{y} \neq y \\
-\epsilon_{acc}, & \text{if prediction is missing}
\end{cases}
\end{equation}
where $\hat{y}$ is the predicted label, $y$ is the ground truth label, and $\epsilon_{acc}$ is a small constant introduced to impose an appropriate penalty when the prediction is missing, thereby encouraging the model to generate correct labels.

\subsubsection{Task-Aware Variable-Length Reward}

To enable the model to flexibly choose between \textbf{long} and \textbf{short} inference modes for each emotion task, we first define an expected length $L_{base}$, a maximum length $L_{max}$, and a minimum length $L_{min}$ for each inference mode within each task. Specifically, we statistically analyze the length distribution of responses under different inference modes within our constructed SFT dataset $\mathcal{D}$. To mitigate the influence of non-normal distributions, we utilize quantile-based statistics derived from the response length distribution $L_l$ to establish the inference length boundaries. Thus, the minimum and maximum lengths are defined as follows:
\begin{equation}
L_{min} = \lfloor P_{5}(L_l) \rfloor, L_{max} = \lceil P_{95}(L_l) \rceil
\end{equation}
where $P_k$ denotes the $k$-th percentile of the distribution. Considering the varying complexity of different emotion tasks, we empirically set an expected length $L_{exp}$ based on domain expertise. Meanwhile, we adopt the statistical median of the length distribution as $L_{sts} = median(L_l)$. Consequently, the final expected length $L_{base}$ is formulated as a weighted combination of empirical and statistical values:
\begin{equation}
L_{base} = \alpha \cdot L_{exp} + (1 - \alpha) \cdot L_{sts}
\end{equation}
where $\alpha$ is an adjustable parameter that balances between empirical knowledge and statistical observations.

Furthermore, we introduce a length-based reward $r_{length}$ to penalize responses deviating from the expected length range and reward responses close to the desired length. Specifically, the length reward is computed as follows:
\begin{equation}
\medmath{
r_{\text{length}} = 
\begin{cases}
\big(s_{\min} \frac{L}{L_{\min}}\big)^{\!2}, & L < L_{\min} \\
\exp\big(s_{\max} \frac{L - L_{\max}}{L_{\max}}\big), & L > L_{\max} \\
\exp\Big(-\frac{1}{2} \big(\frac{L - L_{\text{base}}}{s_{\text{base}}(L_{\max} - L_{\min})}\big)^{\!2}\Big), & L_{\min} \leq L \leq L_{\max}
\end{cases}
}
\end{equation}
Here, $L$ denotes the length of the generated response, while $s_{min}$, $s_{max}$, and $s_{base}$ are scaling factors that control the intensity of rewards and penalties within the reward function. Specifically, for responses shorter than the minimum length, we apply a quadratic penalty to encourage the model to produce more comprehensive inferences. For responses exceeding the maximum length, we employ an exponential penalty to discourage the generation of redundant information. For responses within the acceptable length boundaries, a Gaussian-shaped reward is used to incentivize the model to generate outputs close to the expected length. Through this carefully designed reward mechanism, we expect the model to autonomously adjust its inference length based on the varying difficulty levels of different emotion tasks.

\subsubsection{Reasoning Structure Reward}

During the inference process, whether or not a structured reasoning approach is adopted significantly determines the length and clarity of the reasoning chain generated by the model. To encourage the model to produce responses that demonstrate a clear and structured reasoning process, we propose a novel structure-oriented reward $r_{struct}$. This reward explicitly evaluates key reasoning behaviors (e.g., "decomposition," "reflection," and "verification") and the appropriate usage of logical connectives (e.g., "therefore," "however," and "thereby"). Formally, the proposed reward function is defined as follows:
\begin{equation}
r_{struct} = \lambda \cdot \min\left(\frac{|A|}{N_{A}}, 1\right) + (1-\lambda) \cdot \min\left(\frac{|C|}{N_{C}}, 1\right)
\end{equation}
The set $A$ denotes valid reasoning actions appearing in the response, while $C$ represents the logical connectives used. $N_{A}$ and $N_{C}$ are predefined target numbers of reasoning actions and logical connectives, respectively, and $\lambda$ is a hyperparameter balancing their relative importance. 
Given the variability of optimal reasoning patterns across tasks, our reward design deliberately refrains from constraining the sequence or positioning of reasoning actions and connectives. A response receives the reward if it contains the targeted number of reasoning actions and connectives. To better align with NLP conventions and emphasize meaningful reasoning, we typically set the hyperparameter $\lambda$ higher, placing greater reward emphasis on valid reasoning actions.

\subsubsection{Repetition Penalty}

Additionally, we introduce a repetition penalty $r_{repeat}$, which discourages redundant content within generated responses. The similarity between sentence pairs within the response is determined by combining lexical overlap and semantic similarity from BERT embeddings:
\begin{equation}
Sim(s_i, s_j) = \beta \cdot S_{lex}(s_i, s_j) + (1 - \beta) \cdot S_{sem}(s_i, s_j)
\end{equation}
where lexical similarity $S_{lex}(s_i, s_j)$ is computed using the Jaccard similarity between word sets of the two sentences:
\begin{equation}
S_{lex}(s_i, s_j) = \frac{|W(s_i) \cap W(s_j)|}{|W(s_i) \cup W(s_j)|}
\end{equation}
with $W(s)$ representing the set of words in sentence $s$. Semantic similarity $S_{sem}(s_i, s_j)$ is calculated as the cosine similarity between BERT embeddings of the two sentences:
\begin{equation}
S_{sem}(s_i, s_j) = \frac{\text{BERT}(s_i) \cdot \text{BERT}(s_j)}{|\text{BERT}(s_i)||\text{BERT}(s_j)|}
\end{equation}
The parameter $\beta$ and the similarity threshold $\tau$ are both set empirically based on practical considerations. Specifically, we define the repetition penalty $r_{repeat}$ based on the proportion of sentence pairs whose similarity exceeds the threshold $\tau$: $r_{repeat} = \min\left(\frac{C_\tau}{T},\,1.0\right)$, where $C_\tau$ is the number of sentence pairs with similarity exceeding the empirically chosen threshold $\tau$, and $T$ is the total number of sentence pairs. 

This ensures that responses with higher redundancy receive a greater negative penalty, effectively promoting more concise and diverse generated content.

\subsubsection{Final Reward Function}

The total reward $r_{\text{total}}$ for each response is defined as the weighted sum of four components:
\begin{equation}
\begin{aligned}
r_{\text{total}} ={}& w_{\text{acc}} \cdot r_{\text{acc}} + w_{\text{length}} \cdot r_{\text{length}} \\
& + w_{\text{struct}} \cdot r_{\text{struct}} - w_{\text{repeat}} \cdot r_{\text{repeat}}
\end{aligned}
\end{equation}
We assign the highest weight to $w_{\text{acc}}$ to ensure prediction accuracy remains the top priority. For Long-CoT reasoning, $w_{\text{struct}}$ receives the second-highest weight to encourage structured reasoning; for Short-CoT, we set $w_{\text{struct}} = 0$ to disable this term.
To encourage appropriate response lengths, $w_{\text{length}}$ is given the third-highest weight. Finally, $w_{\text{repeat}}$ receives the lowest weight, applying mild penalties to promote output diversity without sacrificing accuracy.

These weights are empirically tuned through extensive experimentation to effectively balance accuracy, structure, length control, and content diversity.

\begin{table*}[t]
\centering

\renewcommand{\arraystretch}{1.0} % 减少行间距
\setlength{\tabcolsep}{4pt} % 减少列间距

\begin{minipage}{0.48\textwidth}
\centering
\textbf{(a) Sentiment}
\begin{tabular}{lccc}
\hline
Model & Acc & Macro-F1 & Weighted-F1 \\
\hline
LLaMA-3.1-8B & 0.6073 & 0.5391 & 0.5835 \\
DeepSeek-V3 & 0.6314 & 0.6148 & 0.6316 \\
GPT-4o & 0.6326 & 0.5924 & 0.6196 \\
GLM-4 & 0.6326 & 0.6106 & 0.6326 \\
DeepSeek-R1 & 0.6402 & 0.6134 & 0.6381 \\
Qwen-2.5-7B & 0.6437 & 0.6086 & 0.6381 \\
Claude-3.7 & 0.6483 & 0.6184 & 0.6408 \\
Qwen-3-14B & 0.6494 & 0.6007 & 0.6331 \\
Grok-3 & 0.6559 & 0.6295 & 0.6518 \\
\textbf{Emotion-o1} & \textbf{0.6613} & \textbf{0.6405} & \textbf{0.6611} \\
\hline
\end{tabular}
\end{minipage}
\hspace{1em} % 更紧凑的水平间距
\begin{minipage}{0.48\textwidth}
\centering
\textbf{(b) Emotion}
\begin{tabular}{lccc}
\hline
Model & Acc & Macro-F1 & Weighted-F1 \\
\hline
Qwen-2.5-7B & 0.5586 & 0.4158 & 0.5700 \\
LLaMA-3.1-8B & 0.5613 & 0.3206 & 0.5047 \\
DeepSeek-V3 & 0.5801 & 0.4783 & 0.5878 \\
DeepSeek-R1 & 0.5900 & 0.4697 & 0.5967 \\
Qwen-3-14B & 0.5966 & 0.4261 & 0.5811 \\
GPT-4o & 0.6015 & 0.4499 & 0.5812 \\
\textbf{Emotion-o1} & \textbf{0.6107} & \textbf{0.4658} & \textbf{0.6063} \\
GLM-4 & 0.6215 & 0.4819 & 0.6146 \\
Grok-3 & 0.6261 & 0.5058 & 0.6265 \\
Claude-3.7 & 0.6310 & 0.4814 & 0.6098 \\
\hline
\end{tabular}
\end{minipage}

\vspace{0.5ex} % 更紧凑的垂直间距

\begin{minipage}{0.48\textwidth}
\centering
\textbf{(c) Humor}
\begin{tabular}{lccc}
\hline
Model & Acc & Macro-F1 & Weighted-F1 \\
\hline
LLaMA-3.1-8B & 0.6270 & 0.5876 & 0.5871 \\
Qwen-2.5-7B & 0.6475 & 0.6092 & 0.6097 \\
Qwen-3-14B & 0.7373 & 0.7233 & 0.7236 \\
\textbf{Emotion-o1} & \textbf{0.7694} & \textbf{0.7684} & \textbf{0.7684} \\
DeepSeek-R1 & 0.8342 & 0.8306 & 0.8307 \\
DeepSeek-V3 & 0.8506 & 0.8481 & 0.8482 \\
Grok-3 & 0.8869 & 0.8858 & 0.8859 \\
GLM-4 & 0.8989 & 0.8989 & 0.8989 \\
GPT-4o & 0.9043 & 0.9041 & 0.9041 \\
Claude-3.7 & 0.9488 & 0.9488 & 0.9488 \\
\hline
\end{tabular}
\end{minipage}
\hspace{1em}
\begin{minipage}{0.48\textwidth}
\centering
\textbf{(d) Sarcasm}
\begin{tabular}{lccc}
\hline
Model & Acc & Macro-F1 & Weighted-F1 \\
\hline
Qwen-2.5-7B & 0.5030 & 0.3460 & 0.3437 \\
LLaMA-3.1-8B & 0.5360 & 0.4714 & 0.4701 \\
DeepSeek-V3 & 0.6802 & 0.6518 & 0.6511 \\
DeepSeek-R1 & 0.7062 & 0.6947 & 0.6943 \\
Qwen-3-14B & 0.7084 & 0.6895 & 0.6890 \\
GLM-4 & 0.7247 & 0.7174 & 0.7171 \\
Grok-3 & 0.7257 & 0.7126 & 0.7121 \\
Claude-3.7 & 0.7328 & 0.7267 & 0.7270 \\
\textbf{Emotion-o1} & \textbf{0.7469} & \textbf{0.7468} & \textbf{0.7469} \\
GPT-4o & 0.7577 & 0.7570 & 0.7569 \\
\hline
\end{tabular}
\end{minipage}
\caption{Model performance comparison across four emotion recognition tasks}
\label{tab:classification}
\end{table*}
\section{Experiments}
% All papers submitted for publication by AAAI Press must be accompanied by a valid signed copyright form. They must also contain the AAAI copyright notice at the bottom of the first page of the paper. There are no exceptions to these requirements. If you fail to provide us with a signed copyright form or disable the copyright notice, we will be unable to publish your paper. There are \textbf{no exceptions} to this policy. You will find a PDF version of the AAAI copyright form in the AAAI AuthorKit. Please see the specific instructions for your conference for submission details.

% \label{sec:others}

% In this section, we present our experimental design, including the Baseline, Training Setup, and Evaluation Setup. Subsequently, we report the results achieved by Emotion-o1 across various evaluation metrics. Furthermore, we demonstrate the effectiveness of the proposed method through comprehensive ablation studies and an analysis of CoT length. Finally, we perform an Error Analysis to identify the limitations of the approach.

\subsubsection{Baseline}
Nine SoTA LLMs are included for comparison, including open-source models (LLaMA-3.1-8B, DeepSeek-V3, DeepSeek-R1, GLM-4, Qwen-2.5-7B, Qwen-3-14B) and closed-source models (GPT-4o, Claude-3.7, Grok-3).

\subsubsection{Training Setup}

Our experiments are conducted on 4×A100 40G GPUs, running on Ubuntu 22.04, with Python 3.12, PyTorch 2.4.0, and CUDA 12.1. We adopt the \textbf{LLaMA-3.1-8B} as our base model. During the SFT stage, we train the model for $3$ epochs using a learning rate of $2e-5$. Subsequently, we perform the RL stage using PPO, training the model for $4$ epochs with a learning rate of $1e-5$. 
% During training, we generate $4$ responses per prompt to ensure diversity and high-quality outputs.

\subsubsection{Evaluation Setup}

We conduct tasks on the test set of the benchmark dataset, then set the temperature parameter $t = 0.0$ for classification and $t = 0.7$ for generation. We use evaluation metrics widely used for classification tasks: Accuracy, Macro-F1, and Weighted-F1.

\subsubsection{Main Result}
Table~\ref{tab:classification} summarizes the performance comparison between Emotion-o1 and nine baseline models across four emotion tasks, using the average results from 3 experiments. Results show that Emotion-o1 consistently outperforms baselines to varying degrees.

Specifically, compared with the backbone model LLaMA-3.1-8B, emotion-o1 achieves accuracy improvements of \textbf{6\%}, \textbf{5\%}, \textbf{14\%}, and \textbf{21\%} on the sentiment, emotion, humor, and sarcasm tasks, respectively. Additionally, significant enhancements are observed in terms of Macro-F1 and Weighted-F1 scores. For simple tasks (sentiment and emotion), the Macro-F1 score increases by up to \textbf{10\%}, while for more complex tasks (humor and sarcasm), the Weighted-F1 score improves by up to \textbf{27\%}. These findings confirm the effectiveness of explicitly modeling and adapting varying reasoning lengths and strategies to diverse emotion understanding tasks.

We further conducted a comparative evaluation of emotion-o1 against several widely recognized LLMs. Experimental results demonstrate that emotion-o1 achieves SoTA performance on the sentiment task, surpassing established baseline models in terms of accuracy, Macro-F1, and Weighted-F1 metrics. Regarding the sarcasm task, emotion-o1 obtains competitive yet slightly suboptimal performance, achieving results marginally just below GPT-4o $1\%$ but outperforming most other baseline methods. 

In emotion recognition, while Emotion-o1 exceeds leading closed-source models GPT-4o, it exhibits a 2\% deficit relative to SOTA Claude-3.7. Similarly, for humor detection, Emotion-o1 achieves parity with larger-scale models like Qwen3-14B, outperforming it by 3\%, yet demonstrates a substantial gap against top competitors such as Claude-3.7 in complex humor tasks.

Overall, these results highlight that our 8B model achieves competitive or superior performance compared to SoTA LLMs across all tasks, indicating strong generalization and room for further gains with larger-scale base model.

\begin{table}[t]
\centering
\renewcommand{\arraystretch}{1.1}
\setlength{\tabcolsep}{4pt}
\begin{tabular}{l|c|c|c}
    \hline
    & \textbf{Base} & \textbf{SFT} & \textbf{RL} \\
    \hline
    Sentiment & 114 & 93 \textbf{(\textcolor{green}{↓}18\%)} & 85 \textbf{(\textcolor{green}{↓}25\%)} \\
    Emotion & 123 & 186 \textbf{(\textcolor{red}{↑}51\%)} & 183 \textbf{(\textcolor{red}{↑}49\%)} \\
    Sarcasm & 52 & 198 \textbf{(\textcolor{red}{↑}281\%)} & 199 \textbf{(\textcolor{red}{↑}283\%)} \\
    Humor & 70 & 242 \textbf{(\textcolor{red}{↑}246\%)} & 233 \textbf{(\textcolor{red}{↑}233\%)} \\
    \hline
\end{tabular}
\caption{The average CoT length of the model in each stage. \textcolor{green}{↓} indicates shorter, \textcolor{red}{↑} indicates longer.}
\label{tab:stage_ablation}
\end{table}

\begin{figure}[t]
    \centering
    \includegraphics[width=1.0\columnwidth]{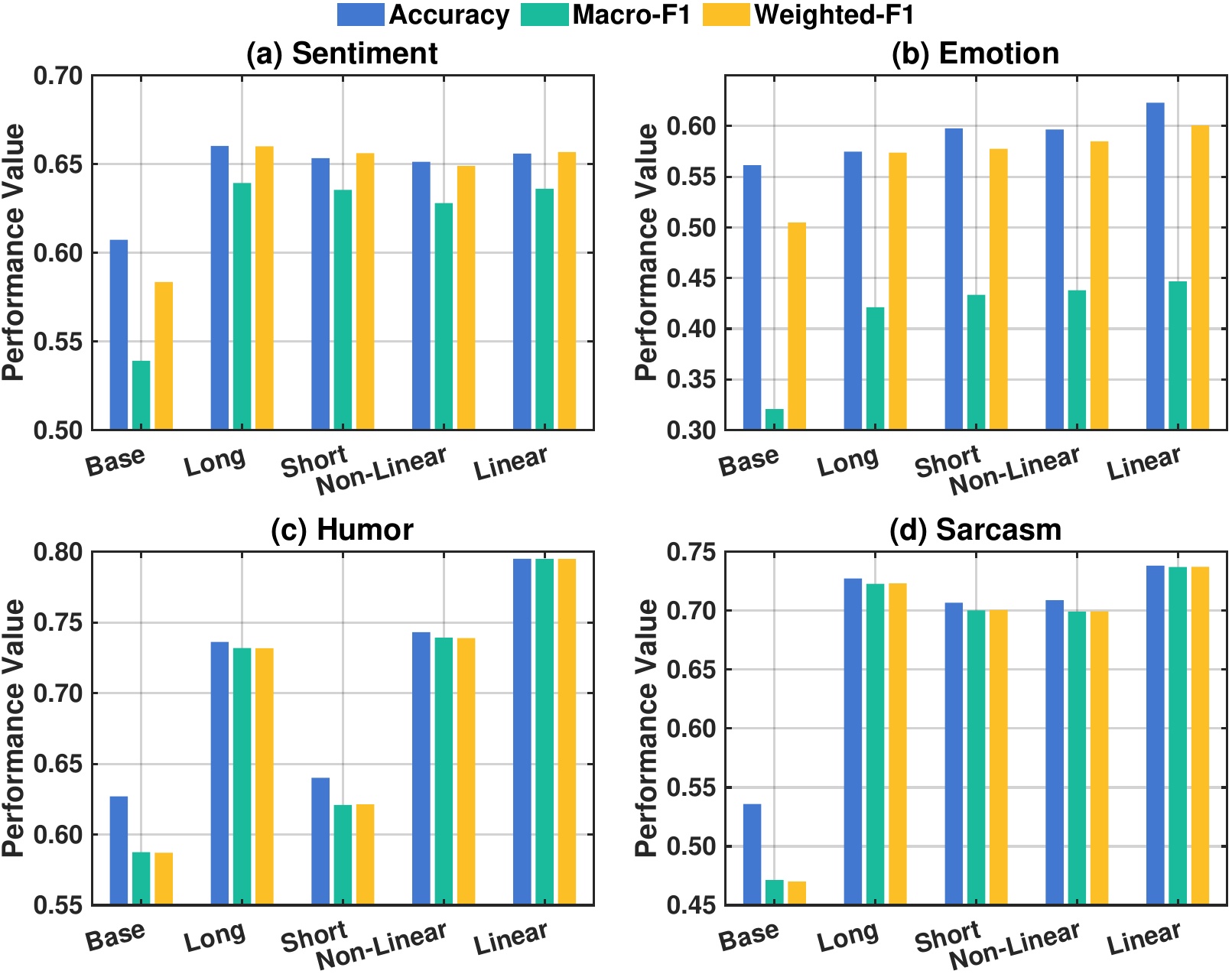}
    \caption{Ablation analysis of CoT variants: Long, Short, Linear and Non-Linear Reasoning}
    \label{fig:reasoning_ablation}
\end{figure}

\subsubsection{CoT Length Analysis}
We analyze the impact of each training stage on the average length of CoT reasoning, as shown in Table~\ref{tab:stage_ablation}. The base model LLaMA-3.1-8B produces shallow rationales with limited variation across tasks. In contrast, SFT introduces task-specific adaptation: it shortens reasoning for simple tasks like Sentiment (↓18\%), while substantially increasing CoT depth for complex tasks such as Sarcasm (↑281\%) and Humor (↑246\%).
The RL stage further refines this adaptation. It further reduces CoT length for Sentiment and Humor by \textbf{7\%} and \textbf{13\%}, respectively, improving efficiency without degrading performance. 

These results confirm that our two-stage pipeline enables Emotion-o1 to learn task-aware reasoning strategies, by using shorter CoTs for straightforward tasks and longer ones for cognitively demanding tasks, and improves both performance and inference efficiency.

\subsubsection{Ablation Study}
We conduct ablation experiments to examine the effects of CoT length and structure, as illustrated in Fig.~\ref{fig:reasoning_ablation}. The key findings are summarized as follows:

\begin{itemize}
\item \textbf{Sentiment}: All CoT variants surpass the base model, achieving gains of around \textbf{5\%} in accuracy and \textbf{10\%} in F1. Notably, short CoT performs on par with long CoT, suggesting concise reasoning is sufficient for such tasks, offering both effectiveness and computational efficiency.

\item \textbf{Emotion}: Short CoT achieves \textbf{2\%} higher accuracy than long CoT, and linear reasoning outperforms nonlinear reasoning by \textbf{3\%}. This suggests that excessive reasoning may introduce unnecessary complexity or noise, and that direct, focused reasoning is more beneficial.

\item \textbf{Humor}: Long CoT yields an \textbf{11\%} improvement in F1 over short CoT, while linear reasoning provides a \textbf{5\%} gain in accuracy. This confirms the need for deeper and more structured reasoning in humor understanding, where implicit intent and abstract cues are involved.

\item \textbf{Sarcasm}: Similar to humor, the combination of long and linear CoT delivers the best results, highlighting the importance of detailed, explicit reasoning in handling pragmatic and context-dependent phenomena.
\end{itemize}

Overall, all CoT-based strategies clearly improve performance over the base model. Moreover, the results validate that sentiment and emotion tasks benefit from concise and direct reasoning, while sarcasm and humor tasks demand longer and deeper reasoning to capture subtle linguistic cues.

\begin{figure}[t]
    \centering
    \includegraphics[width=1.0\columnwidth]{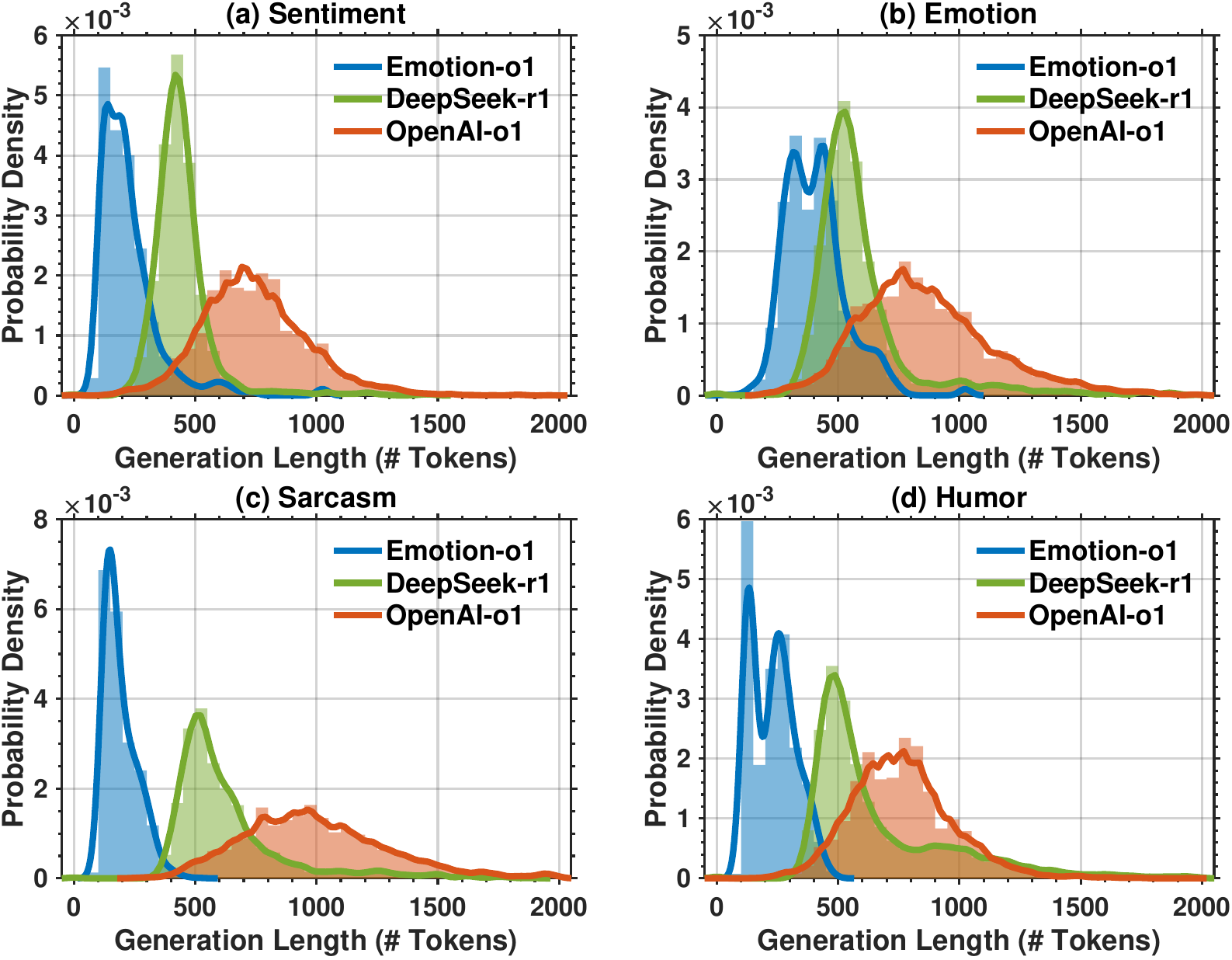}
    \caption{CoT length distribution across reasoning models}
    \label{fig:token_distribution}
\end{figure}

% \begin{figure}[t]
%     \centering
%     \includegraphics[width=0.8\columnwidth]{AnonymousSubmission/LaTeX/error_rate_analysis_academic.pdf}
%     \caption{The average error rate of the four tasks.}
%     \label{fig:error_rate}
% \end{figure}

\subsubsection{Visualization of CoT Length}

We conduct a comparative analysis of Emotion-o1's CoT lengths against the performance of mainstream reasoning models, as illustrated in Figure \ref{fig:token_distribution}. While achieving superior results, Emotion-o1 also demonstrates higher efficiency across all tasks compared to DeepSeek-R1 and OpenAI-o1. Specifically:  

The median length of Emotion-o1 consistently ranks as the lowest among the models. This trend is particularly pronounced in the Sarcasm task, where the length reduces by \textbf{83\%} compared to OpenAI-o1. Even in the Emotion task, which exhibits the smallest improvement, the length decreases by approximately \textbf{52\%} relative to OpenAI-o1.  

Furthermore, Emotion-o1 demonstrates a lower maximum length compared to its competitors. In the Sarcasm task, the 90th percentile length reduces by nearly \textbf{4.8} times relative to OpenAI-o1, underscoring the model's ability to effectively suppress extreme values.  

The standard deviation of lengths generated by Emotion-o1 across tasks is consistently lower. For example, in the Sarcasm task, the standard deviation is only \textbf{1/3} that of OpenAI-o1, indicating Emotion-o1 produces more stable outputs with greater concentration in response values.

In summary, Emotion-o1 consistently exhibits superior output efficiency across all tasks while maintaining competitiveness against mainstream reasoning models.  

\begin{figure}[t]
    \centering
    \includegraphics[width=0.9\columnwidth]{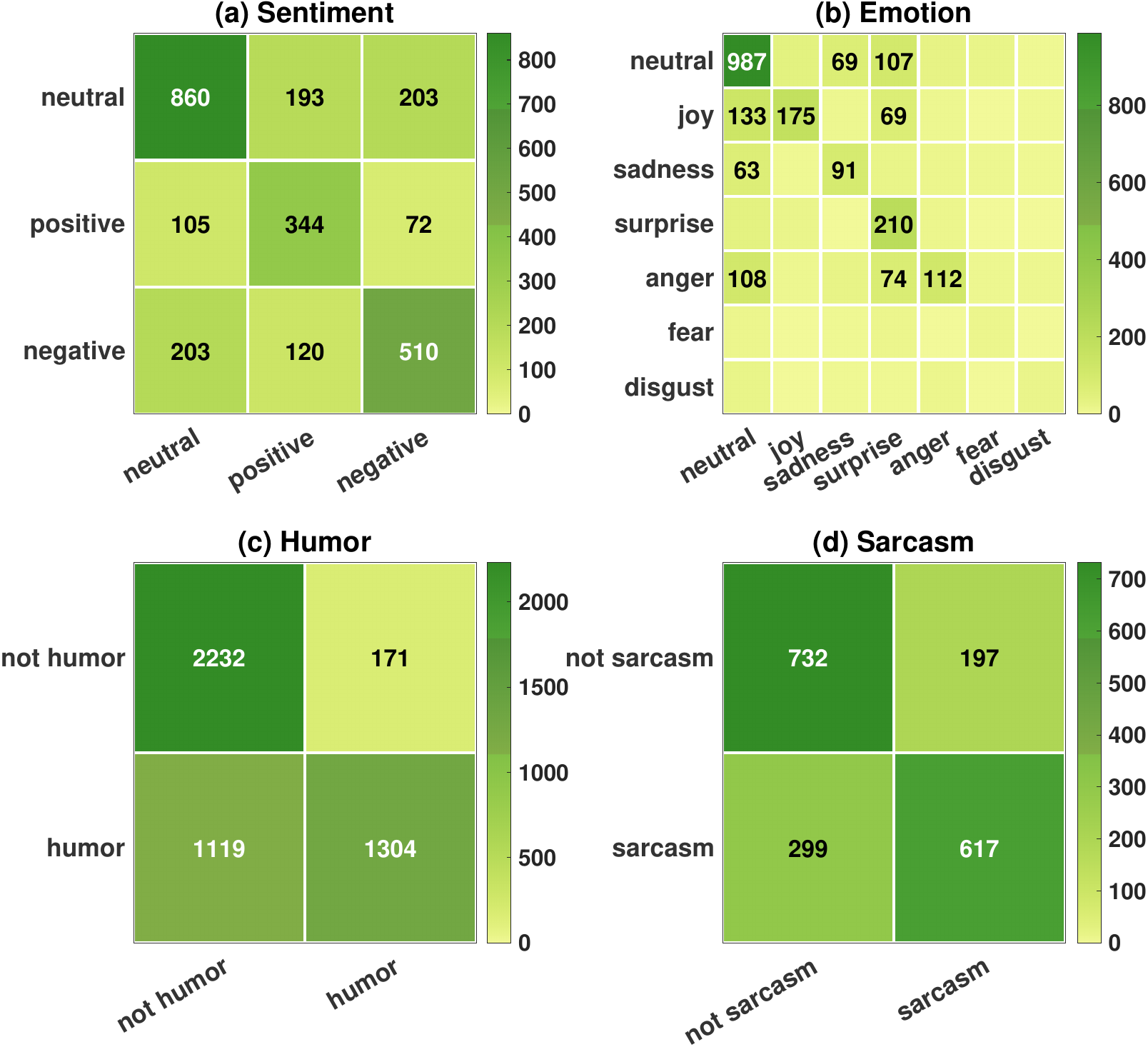}
    \caption{The confusion matrices of the four tasks.}
    \label{fig:confusion_matrices}
\end{figure}

\subsubsection{Error Analysis}

Figure \ref{fig:confusion_matrices} illustrates the error analysis for the four tasks. Sarcasm and Humor detection show notable false negatives, indicating difficulties in identifying complex contextual cues. Sentiment analysis exhibits polarity interpretation challenges, particularly in distinguishing neutral from negative expressions. Emotion recognition demonstrates a tendency to default to neutral classifications, with authentic emotions like joy and anger frequently misclassified, alongside cross-category confusion. The observed false positives in sentiment and sarcasm tasks suggest occasional oversensitivity to certain linguistic signals. Collectively, these patterns highlight opportunities for refining contextual understanding and developing task-specific approaches to capture linguistic nuances more effectively.

\section{Conclusion}

This work addresses the limitations of fixed-length chain-of-thought reasoning for emotional understanding in LLMs by proposing Emotion-o1, an adaptive framework that dynamically adjusts reasoning depth based on task complexity. Through multi-stage training with a multi-objective reward (accuracy, brevity, structure, redundancy), our approach achieves significant performance gains: F1 improvements of 10\% (sentiment), 5\% (emotion), 18\% (humor), and 27\% (sarcasm) over its backbone. Notably, the 8B model outperforms Grok-3 by 1.1\% and Claude-3.7 by 2\% in critical tasks while reducing reasoning length by 83\% versus OpenAI-o1. Emotion-o1 establishes an efficient bridge between structured reasoning and emotional understanding.
\subsubsection{Limitations} 
First, our method is evaluated on four curated emotion-related tasks, which, while diverse, may not cover the full spectrum of affective reasoning challenges in real-world applications. Second, our framework focuses solely on textual input, excluding multimodal signals (e.g., visual or acoustic cues), which are often crucial for understanding emotions in human communication.

\bibliography{aaai2026}
\end{document}